\documentclass[10pt,twocolumn,letterpaper]{article}

\usepackage{iccv}
\usepackage{times}
\usepackage{epsfig}
\usepackage{graphicx}
\usepackage{amsmath}
\usepackage{amssymb}
\usepackage{makecell}
\usepackage{pifont}

\usepackage[accsupp]{axessibility}

\usepackage[dvipsnames,table]{xcolor}
\usepackage{nicefrac}
\usepackage{enumitem}
\usepackage{booktabs}
\usepackage{siunitx}

\definecolor{lightgray}{gray}{0.95}
\definecolor{midgray}{gray}{0.55}
\definecolor{steelblue}{HTML}{4D82B7}
\definecolor{davysgrey}{rgb}{0.33, 0.33, 0.33}

\usepackage[first=0, last=9]{lcg}

\newcommand{\hotagain}[1]{\gainped{#1}}
\newcommand{\idfgain}[1]{\gainped{#1}}

\newcommand{\gainped}[1]{\textcolor{ForestGreen}{$\mathbf{+ \ensuremath{#1}}$}}

\newcommand{\cmark}{\ding{51}}%
\newcommand{\xmark}{\ding{55}}%
\newcommand{\quotationmarks}[1]{``#1''}
\newcommand{\tit}[1]{\smallbreak\noindent\textbf{#1 }}

\newcommand{\tinytit}[1]{\noindent\textbf{#1 }}
\newcommand{\methnam}{TrackFlow\xspace}%
\newcommand{\liftername}{DistSynth\xspace}
\usepackage[pagebackref=true,breaklinks=true,colorlinks,bookmarks=false]{hyperref}

\iccvfinalcopy

\def\iccvPaperID{12578}

\ificcvfinal\fi
\usepackage[capitalize]{cleveref}
\crefname{section}{Sec.}{Secs.}
\Crefname{section}{Section}{Sections}
\Crefname{table}{Table}{Tables}
\crefname{table}{Tab.}{Tabs.}

\begin{document}

\title{TrackFlow: Multi-Object Tracking with Normalizing Flows}

\author{
Gianluca Mancusi$^1$ \qquad Aniello Panariello$^1$ \qquad Angelo Porrello $^1$ \qquad Matteo Fabbri$^2$ \\ Simone Calderara$^1$ \qquad Rita Cucchiara$^{1,3}$\\\\
\begin{tabular}{ccc}
\makecell{$^1$University of Modena and Reggio Emilia, Italy} & \makecell{$^2$GoatAI S.r.l.}&
\makecell{$^3$IIT-CNR, Italy}\\
\end{tabular}
}

\maketitle
\ificcvfinal\thispagestyle{empty}\fi

\begin{abstract}
The field of multi-object tracking has recently seen a renewed interest in the good old schema of \textit{tracking-by-detection}, as its simplicity and strong priors spare it from the complex design and painful babysitting of \textit{tracking-by-attention} approaches. In view of this, we aim at extending tracking-by-detection to multi-modal settings, where a comprehensive cost has to be computed from heterogeneous information \eg, 2D motion cues, visual appearance, and pose estimates. More precisely, we follow a case study where a rough estimate of 3D information is also available and must be merged with other traditional metrics (\eg, the IoU). To achieve that, recent approaches resort to either simple rules or complex heuristics to balance the contribution of each cost. However, \emph{i)} they require careful tuning of tailored hyperparameters on a hold-out set, and \emph{ii)} they imply these costs to be independent, which does not hold in reality. We address these issues by building upon an elegant probabilistic formulation, which considers the cost of a candidate association as the \emph{negative log-likelihood} yielded by a deep density estimator, trained to model the conditional joint probability distribution of correct associations. Our experiments, conducted on both simulated and real benchmarks, show that our approach consistently enhances the performance of several \textit{tracking-by-detection} algorithms.
\end{abstract}
\section{Introduction}
Real-time multi-person tracking in crowded real-world scenarios is a challenging and difficult problem with applications ranging from autonomous driving to visual surveillance. Indeed, the work done to create a reliable tracker that can function in every environment is noteworthy.

The most successful methods currently available in literature can be broadly grouped into three main categories: \textit{tracking-by-detection}~\cite{bochinski2017high,berclaz2011multiple,kim2015multiple}, \textit{tracking-by-regression}~\cite{feichtenhofer2017detect,held2016learning,liu2020gsm}, and \textit{tracking-by-attention}~\cite{meinhardt2022trackformer,yan2021learning,wu2021track}. In \textit{tracking-by-detection}, bounding boxes are computed independently for each frame and associated with tracks in subsequent steps. \textit{Tracking-by-regression} unifies detection and motion analysis, with a single module that simultaneously locates the bounding boxes and their displacement \wrt the previous frame. Finally, in \textit{tracking-by-attention}, an end-to-end deep tracker based on self-attention~\cite{vaswani2017attention} manages the life-cycle of a set of track predictions through the video sequence. 

Although the two latter paradigms have recently sparked the research interest, \textit{tracking-by-detection} still proves to be competitive~\cite{zhang2022bytetrack,cao2022observation}, under its simplicity, reliability, and the emergence of super-accurate object detectors~\cite{ge2021yolox}. In light of these considerations, we aim to strengthen~\textit{tracking-by-detection} algorithms by enriching the information they usually leverage -- \ie, the displacement between estimated and actual bounding boxes~\cite{wojke2017simple,zhang2022bytetrack} -- with additional cues. Indeed, as shown by several works of multi-modal tracking~\cite{chiu2020probabilistic,zhang2019robust}, the visual domain is just one of the possible sources that may contribute. The pose of the skeleton~\cite{clark2019three}, the depth maps~\cite{rajasegaran2022tracking,dendorfer2022quo} and even thermal measurements~\cite{kumar2014improving} are concepts that can gain further robustness, as they encode a deeper understanding of the scene. In particular, as humans move and interact in a three-dimensional space, one of the goals of this work is to provide the tracker with the (predicted) distance from the camera, thus resembling what is generally acknowledged as \quotationmarks{2.5D}. To achieve that, we train a per-istance distance deep regressor on MOTSynth~\cite{fabbri2021motsynth}, a recently released synthetic dataset displaying immense variety in scenes, lightning/weather conditions, pedestrians' appearance, and behaviors.

However, the fusion of multi-modal representations poses a big question: how to weigh the contribution of each input domain to the overall cost? It represents a crucial step, as its design directly impacts the subsequent assignment optimization problem: in this respect, existing works resort to handwritten formulas and heuristics \eg, DeepSORT~\cite{wojke2017simple} computes two different cost matrices and combines them through a weighted sum. Notably, the authors of~\cite{rajasegaran2022tracking} build upon a probabilistic formulation, which recasts the cost $c_{i,j}$ as the likelihood of the event \quotationmarks{the $i$-th detection belongs to the $j$-th tracklet}. Afterward, it is about estimating a density function on top of correct associations, termed \textit{inliers}. Although these fusing approaches may appear reasonable, they hide several practical and conceptual pitfalls:
\begin{itemize}[leftmargin=5.5mm]
\setlength\itemsep{0em}
    \item They introduce additional hyperparameters, which require careful tuning on a separate validation set and hence additional labeled data.
    \item A single choice of these hyperparameters cannot fit different scenes perfectly, as these typically display different dynamics in terms of pedestrians' motion and spatial density, the camera's position/motion, and lighting/weather conditions. Therefore, the right trade-off is likely to be scenario-dependent;
    \item Common approaches (\eg, a simple weighted summation) assume the input modalities to be independent, thus overlooking their interactions.
\end{itemize}
We propose to take into account the weaknesses mentioned above through a dedicated parametric density estimator -- termed \methnam{} -- tasked to summarize several input costs/displacements in a single output metric, \eg, the probability that a specific detection $D$ belongs to a particular track $T$. As we strive to approximate the underlying conditional probability distribution $\mathcal{P}(D \in T | \ T)$ over the input costs, we borrow the estimator from the world of deep generative models, in particular from the literature of Normalizing Flow models~\cite{dinh2014nice,dinh2016density,kingma2018glow}. In fact, these models represent a flexible and effective tool to perform density estimation. Moreover, we would like to emphasize the reliance of such a module on an additional context-level representation, which we provide in order to inform the model about scene-level peculiarities. This way, the computation of the likelihood is also conditioned on visual cues of the scene, which we assume may be unobserved during evaluation. 

Extensive experiments on MOTSynth~\cite{fabbri2021motsynth}, MOT17~\cite{milan2016mot16}, and MOT20~\cite{dendorfer2021motchallenge} show that the naive cost metric -- \ie, the 2D intersection between predicted and candidate bounding boxes -- can be replaced by the score provided by our approach, with a remarkable performance gain in exchange.
\section{Related Works}
\subsection{Multiple object tracking (MOT)}
Since the advent of deep learning, advances in object detection~\cite{redmon2016you,ren2015faster,ge2021yolox,zhou2019objects} drove the community towards \textit{tracking-by-detection}~\cite{bergmann2019tracking,wojke2017simple,bewley2016simple,zhang2022bytetrack,zhou2020tracking,lu2020retinatrack}, where bounding boxes are associated with tracks in subsequent steps. Among the most successful works, Tracktor~\cite{bergmann2019tracking} pushes \textit{tracking-by-detection} to the edge by relying solely on an object detector to perform tracking. CenterTrack~\cite{zhou2020tracking} provides a point-based framework for joint detection and tracking based on CenterNet~\cite{duan2019centernet}. Similarly, RetinaTrack~\cite{lu2020retinatrack} extends RetinaNet~\cite{lin2017focal} to offer a conceptually simple and efficient joint model for detection and tracking, leveraging instance-level embeddings. More recently, ByteTrack~\cite{zhang2022bytetrack} further establishes this paradigm, unleashing the full potential of YOLOX~\cite{ge2021yolox}: notably, it uses almost every predicted detection, and not only the most confident ones.

As Transformers~\cite{vaswani2017attention} gained popularity~\cite{dosovitskiy2020image,he2022masked,liu2021swin,carion2020end}, various attempts have been carried out to apply them to MOT. TransTrack~\cite{sun2020transtrack} leverages the attention-based query-key mechanism to decouple MOT as two sub-tasks \ie, detection and association. Similarly, TrackFormer~\cite{meinhardt2022trackformer} jointly performs tracking and detection, with a single decoder network. Furthermore, MOTR~\cite{zeng2022motr} builds upon DETR~\cite{carion2020end} and introduce \quotationmarks{track queries} to model the tracked instances in the entire video, in an end-to-end fashion. 

Recently, a few attempts have been made to leverage 3D information for MOT. Quo Vadis~\cite{dendorfer2022quo} shows that forecast analysis performed in a bird's-eye view can improve long-term tracking robustness. To do so, it relies on a data-driven heuristics for the homography estimation: hence, it may suffer in presence of non-static cameras or target objects moving on multiple planes. Differently, PHALP~\cite{rajasegaran2022tracking} computes a three-attribute representation for each bounding box \ie, appearance, pose, and location. Similarly to our approach, they adopt a probabilistic formulation to compute the posterior probabilities of every detection belonging to each one of the tracklets.
\tit{MOT and trajectory forecasting}~As our approach computes an approximation of the true $\mathcal{P}(D \in T | \ T)$, it can be thought as a one-step-ahead trajectory predictor, thus resembling a non-linear and stochastic Kalman filter learned directly from data. In this respect, our work actually fits the very recent strand of literature~\cite{saleh2021probabilistic,dendorfer2022quo,kesa2022multiple,monti2022many} that takes into consideration the possible applications of trajectory forecasting in MOT. Similarly to our approach, the authors of~\cite{saleh2021probabilistic} perform density estimation over the location of the bounding box in the next time-step; however, while they rely on PixelRNN~\cite{van2016pixel} and model $\mathcal{P}(D \in T | \ T)$ as a multinomial category distribution, we instead exploit a more flexible and powerful family of generative approaches, such as normalizing flow models~\cite{dinh2014nice,rezende2015variational,dinh2016density}. Differently, Kesa \etal~\cite{kesa2022multiple} propose a deterministic approach that simply regresses the next position of the bounding box, which, arguably, does not consider the stochastic and multi-modal nature of human motion. Finally, the authors of~\cite{monti2022many} show that a teacher-student training strategy helps when only a few past observations are available to the model, as hold in practice for newborn tracklets.
\subsection{Distance estimation}
\begin{figure*}[t]
    \centering
    \includegraphics[width=0.92\linewidth]{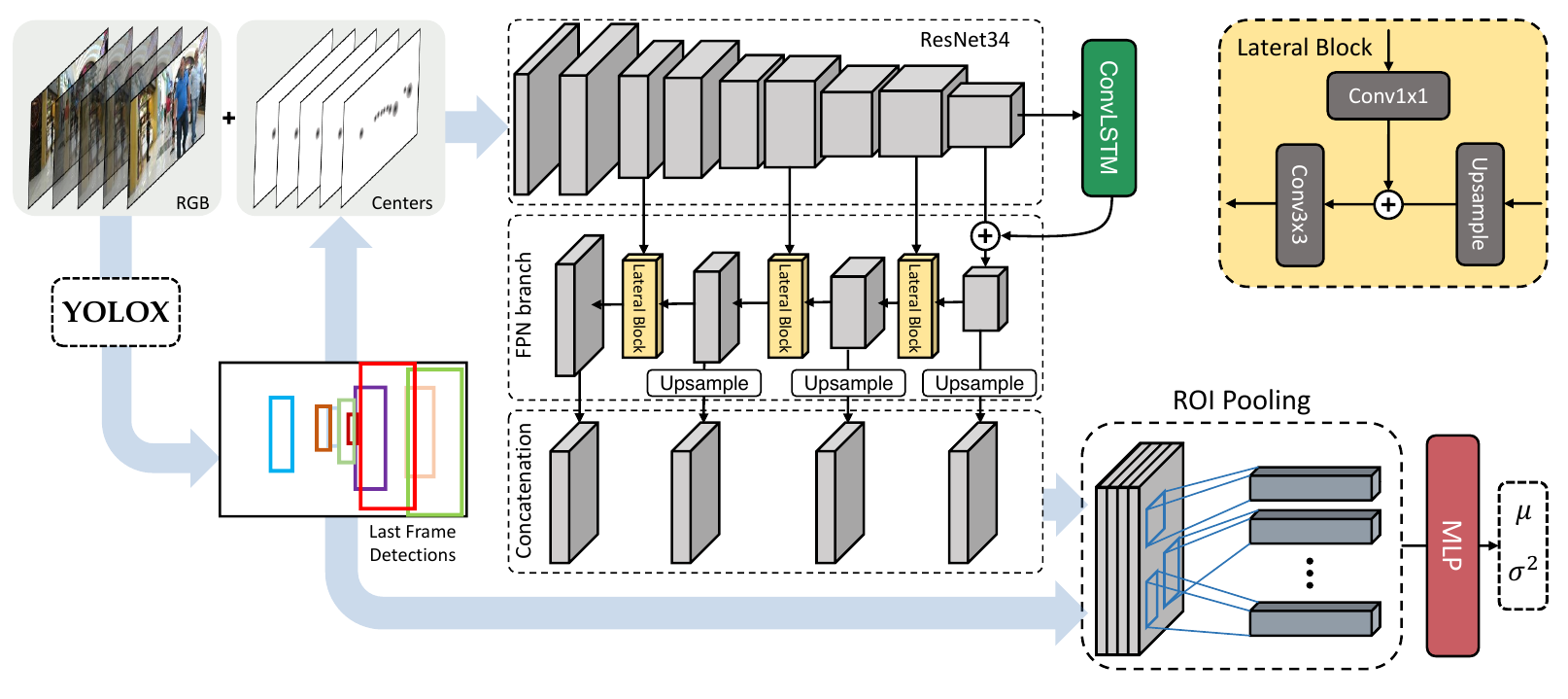}
    \caption{Overview of the camera distance estimator. \liftername predicts per-objects distances from a short video clip. We further provide the centers of each bounding box as an additional input channel. After several convolutional blocks processing each frame independently: \textit{i)} a temporal module is devised to extract temporal patterns; \textit{ii)} the activation maps undergo the FPN branch, in order to preserve local details. Finally, feature maps from distinct layers are stacked and passed to the RoI pooling layer. The latter produces per-pedestrian vector representations, which we finally use to predict the pedestrians' expected distance $\mu$ and uncertainty $\sigma^2$.}
    \label{fig:method}
    \vspace{-0.7em}
\end{figure*}
The estimation of distances from monocular images is a crucial task for many computer vision applications and has been studied for many years~\cite{alhashim2018high,godard2017unsupervised,ranftl2021vision, godard2019digging,ranftl2020towards,lee2019big}. Classical approaches address the problem by regressing the relation between the object's geometry and the distance. Among these, the most popular is the inverse perspective mapping (IPM) algorithm~\cite{mallot1991inverse,rezaei2015robust}, which converts image points into bird's-eye view coordinates in an iterative manner. Unfortunately, due to the presence of notable distortion, IPM may yield unreliable results for objects located far from the camera. Following methods~\cite{haseeb2018disnet, gokcce2015vision} exploit the size of the bounding box to infer the geometry of the object; despite their simplicity and relative effectiveness, these approaches are unreliable if target instances are from different classes (\eg, vehicles, people, \etc), or in the case of high intra-class variance (as hold for the class person).

More recently, Zhu \etal~\cite{zhu2019learning} exploit Deep Convolutional Neural Networks (CNNs) to enrich the representation of each bounding box with visual cues. In details, they firstly extract the detections through Faster R-CNN~\cite{ren2015faster}; afterward, they feed a tailored CNN with the whole frame and finally apply Region of Interest (RoI) pooling, which spatially aggregates activation maps and outputs one representation for each retrieved detection. As discussed in the following section, overlapping objects may lead to erroneous predictions, as the features of foreground objects may contaminate the activations of the occluded ones.
\section{Method}
Our architecture comprises of two main building blocks:
\begin{itemize}[leftmargin=5.5mm]
\setlength\itemsep{0em}
    \item A deep neural regressor that, given a monocular image, estimates the distance of each pedestrian from the camera (see~\cref{sec:distsec}). We called it \textbf{\liftername}, as we train only on synthetic images from MOTSynth~\cite{fabbri2021motsynth}.
    \item A deep density estimator, termed \textbf{\methnam} (\cref{sec:density}), which has to merge 2D cues (\eg, the spatial displacement between bounding boxes) with the 3D localization information obtained through \liftername.
\end{itemize}
\subsection{\liftername: estimating per-instance distance from a monocular image}
\label{sec:distsec}
As the output of the distance estimator is meant to further refine the association cost between detections and tracks, it is crucial to handle possible temporary occlusions and noisy motion patterns. Therefore, as deeply discussed in the following, our model integrates time information (\eg, a short collection of past frames) with visual cues to achieve a smoother and more reliable distance prediction.

In details, the network is fed with a short video clip $\mathbb{R}^{T\times C \times W \times H}$, where $T$ is the length of the video clip, $C$ is the number of channels, $W$ and $H$ are the width and height of the frames. As we are not interested in a dense prediction for the entire scene but in pedestrian-level predictions, we ask the network to focus its attention on a restricted set of locations: namely, in proximity of the bounding boxes provided by an off-the-shelf detector, \eg, YOLOX~\cite{ge2021yolox}. To do so, we concatenate an additional channel to the RGB frames, representing the center of each bounding box.

The architecture mainly follows the design of residual networks~\cite{he2016deep} (in our experiments, we used ResNet-34 pretrained on ImageNet~\cite{deng2009imagenet}). Importantly, we apply two modifications to the feature extractor to enhance its capabilities, discussed in the following two sub-paragraphs.

\tit{Exploiting temporal information}~While related works~\cite{zhu2019learning,haseeb2018disnet} focus solely on the last frame of interest, we propose to condition the predictions of camera distances on a small window of previous frames, thus encompassing temporal dynamics. The main goal is to provide a much more robust prediction when the target pedestrian is partially or temporarily occluded in the current frame but visible in the previous ones. In that case, his/her history would compensate and smooth the prediction. Therefore, we equip the backbone with a layer capable of processing the sequence of past feature maps: precisely, a Convolutional Recurrent Neural Network, \ie, a ConvLSTM~\cite{shi2015convolutional}, whose output is a single-frame feature map encoding all the history of past frames. We insert such a module in the deeper layer of the network \ie, to the exit of the last residual block of our backbone. 

\tit{Improving spatial representations}~Standard CNNs usually exploit pooling layers to progressively down-size the activation maps. For classification, there are few doubts that such an operation provides advantageous properties (\eg, translation invariance, high-level reasoning, \etc). Considered the task of per-object distance estimation, instead, we argue that the reliance on pooling layers could be detrimental. In fact, considered people located far away from the camera (\ie, those surrounded by tiny bounding boxes), pooling layers could over-subsample the corresponding spatial regions, with a significant loss in terms of visual cues. 

To avoid such a detrimental issue, we equip the feature extractor with an additional branch based on Feature Pyramid Network (FPN)~\cite{lin2017feature}. In practice, it begins with the encoding produced by the temporal module, then it proceeds in the reverse direction (\ie, from deepers layer to the one closer to the input), and restores the original resolution through up-sampling layers and residual paths from the forward flow. To gain a deeper understanding, \cref{fig:method} proposes a comprehensive visual of the architecture.

\tit{Output and loss function}~Once the feature maps have been processed through the temporal module and the pyramid, we again exploit the bounding boxes and perform RoI pooling~\cite{girshick2015fast} to obtain a feature vector for each pedestrian. The result is a $\mathbb{R}^{N\times H \times K \times K}$ feature map, where $N$ indicates the number of detected pedestrians, $H$ the number of hidden channels, and $K = 4$ the dimension of the RoI pooling window. We process these feature maps through a multilayer perceptron (MLP), which outputs the predicted distances. Finally, we do not make a punctual estimate but ask the network to place a Gaussian distribution over the expected distance, thus obtaining the model's aleatoric uncertainty~\cite{bertoni2019monoloco,der2009aleatory}. In practice, it translates into yielding two values, $d \equiv d_{\mu}$ and $d_{\sigma^2}$, and optimizing the Gaussian Negative Log Likelihood (GNLL)~\cite{nix1994estimating}, as follows:
\begin{equation*}
\label{eq:gnll}
\operatorname{GNLL}(d_{true} \lvert d, d_{\sigma^2})=\frac{1}{2} \left( \log(d_{\sigma^2}) + \frac{(d - d_{true})^2}{d_{\sigma^2}}\right).
\end{equation*}
\subsection{TrackFlow: modeling the density of correct associations through Normalizing Flows}\label{sec:density}
\subsubsection{Problem statement}
In a nutshell, the \textit{tracking-by-detection} paradigm usually relies on the Kalman filter~\cite{smeulders2013visual,bewley2016simple} to estimate the next 2D spatial position $\mathbf{p}^{t+1}_{j} = [x^{t+1}_{j}, y^{t+1}_{j}]$ of a certain pedestrian $j$ in the next $t+1$-th frame. The prediction $\hat{\mathbf{p}}^{t+1}_{j}$ depends upon the set of previous observations, contained in a short track $T_j = [\mathbf{p}^{t}_{j}, \mathbf{p}^{t-1}_{j}, \dots, \mathbf{p}^{t-\lvert T \lvert+1}_{j}]$ recording the past matched locations of the pedestrian $j$. Afterward, given a new set of detections $D_i = [\mathbf{p}_{i}, \mathbf{w}_{i}, \mathbf{h}_{i}] \ \ i=1,2,\dots,\lvert D \lvert$ (with $\mathbf{w}_{i}$ and $\mathbf{h}_{i}$ being the width and the height of the bounding box respectively), the \textit{cost} of a candidate association between $D_i$ and the track $T_j$ can be computed as the displacement $\Delta_p \equiv \Delta_p(T_j, D_i)$ between the predicted location and the candidate one, \ie, $\Delta_p = \operatorname{d}(\hat{\mathbf{p}}^{t+1}_{j}, \mathbf{p}_{i})$. In such a notation, $\operatorname{d}(\cdot, \cdot)$ stands for any function penalizing such a displacement, as the Euclidean distance $||\hat{\mathbf{p}}^{t+1}_{j} - \mathbf{p}_{i}||_2^2$. Similarly, the variation of the sizes of the bounding box, \ie, $\Delta_{w,h} \equiv \Delta_{w,h}(T_j, D_i)$ could be taken into account.

Furthermore, by virtue of the regressor introduced in~\cref{sec:distsec}, we could additionally exploit the displacement beliefs/reality relating to camera distances $\Delta_{d} = \operatorname{d}(\hat{d}^{\ t+1}_{j}, d_{i})$, given the estimated one-step-ahead distance $\hat{d}^{\ t+1}_{j}$ for the track $T_j$ and the distance $d_{i}$ of the candidate detection $D_i$, inferred through \liftername. To ease the notation, from now on we will denote $T_j$ / $D_i$ as $T$ / $D$.

Once we have computed these costs (but other could be profitably envisioned), we shall define an aggregated cost function $\Phi(T, D) = f(\Delta_p,\Delta_{w,h},\Delta_d)$ that jointly computes the cost of the candidate association $D \in T$. There are several approaches to achieve that; among those, we build upon the probabilistic formulation proposed in~\cite{rajasegaran2022tracking} and define the cost $\Phi$ as negative log-(conditional) likelihood:
\begin{align*}
\Phi(T, D) &= -\log \mathcal{P}_{\theta}(D \in T \ | \ T)\\
&= -\log f([\Delta_p,\Delta_{w,h},\Delta_d] \ | \ T, \theta).
\end{align*}
In that formulation, the target conditional probability distribution $\mathcal{P}_{\theta}(\cdot)$ parametrizes as a learnable function $f(\cdot \ | \ \theta)$, where its parameters $\theta$ have to be sought by maximizing the likelihood of correct associations (often referred to as \textit{inliers}). To ease the optimization, the authors of~\cite{rajasegaran2022tracking} factorized the above-mentioned density, assuming that each of the marginal distributions is independent, such that $\mathcal{P}_{\theta}(D \in T \ | \ T) \propto \mathcal{P}_p \ \mathcal{P}_{w,h} \ \mathcal{P}_d$. Therefore:
\begin{equation*}
\Phi(T, D)=-\log\mathcal{P}_{\theta_1}(\Delta_p)- \log\mathcal{P}_{\theta_2}(\Delta_{w,h})-\log\mathcal{P}_{\theta_3}(\Delta_{d}).
\end{equation*}
As discussed in the next subsection, we do not impose such an assumption but approximate, via Maximum Likelihood Estimation (MLE), the joint conditional distribution with a deep generative model $f(\cdot \ | \ T, \theta)$.
\subsubsection{Overview of the architecture}
Among many possible choices (\eg, variational autoencoders~\cite{kingma2013auto}, generative adversarial networks~\cite{goodfellow2020generative} or the most recent diffusion models~\cite{ho2020denoising}), we borrow the design of $f(\cdot|T, \theta)$ from the family of normalizing flow models~\cite{dinh2014nice,rezende2015variational,dinh2016density}. Notably, they provide an exact estimate of the likelihood of a sample, in contrast with other approaches that yield a lower bound (as the variational autoencoder and its variants~\cite{van2017neural,tomczak2018vae}). Moreover, normalizing flow models grant high flexibility, as they do not rely on a specific approximating family for the posterior distribution. The latter is instead a peculiar trait of the variational methodology, which may suffer if the approximating family does not contain the true posterior distribution.

Briefly, a normalizing flow model creates an invertible mapping between a simple factorized base distribution with known density (\eg, a standard Gaussian in our experiments) and an arbitrary, complex and multi-modal distribution, which in our formulation is the conditional distribution $\mathcal{P}(D \in T \ | \ T)$ underlying correct associations. The mapping between the two distributions is carried out through a sequence of $L$ invertible and differentiable transformations $g_l(\cdot\ | \ T)$ (with parameters $\theta_l$, omitted in the following), which progressively refines the initial density through the rule for change of variables. In formal terms, our proposal named \textbf{\methnam} takes the following abstract form:
\begin{equation}
\label{eq:trackflow}
f([\Delta_p,\Delta_{w,h},\Delta_d] \ | \ T, \theta) = g^{-1}_L \circ \dots \circ g^{-1}_2 \circ g^{-1}_1, 
\end{equation}
where
\begin{eqnarray*}
\operatorname{forward \ pass}:& \mathbf{z}_{l} = g_l(\mathbf{z}_{l-1} \ | \ T); \ \mathbf{z}_{L} \sim \mathcal{P}_{\theta}(D \in T \ | \ T) \\
\operatorname{inverse \ pass}:& \mathbf{z}_{l-1} = g^{-1}_l(\mathbf{z}_{l} \ | \ T); \ \mathbf{z}_{0} \sim \mathcal{N}(0,\,1)
\end{eqnarray*} 
are the forward pass (\ie, used when sampling) and the inverse pass (\ie, used to evaluate densities) of \methnam. The model can be learned via Stochastic Gradient Descent (SGD), by minimizing the negative log-likelihood on a batch of associations sampled from the true $\mathcal{P}(D \in T \ | \ T)$ (\ie, corresponding to valid tracks). The loss function exploits the inverse pass and takes into account the likelihood under the base distribution~\cite{kobyzev2020normalizing} plus an additive term for each change of variable occurred through the flow.

\tit{Base architecture} Regarding the design of each layer $g_l(\cdot\ | \ T)$, we make use of several well-established building blocks, such as normalization layers, masked autoregressive layers~\cite{papamakarios2017masked}, and invertible residual blocks~\cite{chen2019residual}. In particular, our model features a cascade of residual flows~\cite{behrmann2021understanding}, which we preferred to other valuable alternatives (\eg, RealNVP~\cite{dinh2016density}) in light of their expressiveness and proven numerical stability. For the sake of conciseness we are omitting the inverse functions, but the overall representation of the forward pass of the $l$-th block proceeds as follows:
\begin{eqnarray*}
    \operatorname{residual \ block}:& z = \operatorname{MLP}_l(\mathbf{z}_{l-1}) + \mathbf{z}_{l-1},\ \\
    \operatorname{act. \ norm}:& z = s_{l} \odot z + b_{l}, \\
    \operatorname{masked \ auto. \ flow}:& \mathbf{z}_{l} = \operatorname{MAF}_l(\operatorname{concat}[z \ \lvert\lvert \ \operatorname{e}_l]),
\end{eqnarray*}
where $\operatorname{e}_l$ refers to an auxiliary learnable representation discussed below, by which we take into account the dependence on the external context (\eg, the track $T$).
\subsubsection{Context encoder}
\label{sec:contextencoder}
\tit{Dependence on temporal cues} As stated by Eq.~\ref{eq:trackflow}, the inverse pass (but also the forward one) of \methnam depends also on the observed track $T$. By introducing such a conditioning information, the model could learn to assign higher likelihood to the candidate associations that exhibit motion patterns coherent with those observed in the recent past. To introduce such an information, we take inspiration from~\cite{winkler2019learning,scholler2021flomo} and condition each invertible layer on an additional latent representation $e_l$. The latter is given by a tailored temporal encoder network $\operatorname{e}_{\theta_l} \text{s.t.} \ e_l = \operatorname{e}_{\theta_l}(T)$ fed with the observed track $T$.

Importantly: \textit{i)} as advocated in several recent works~\cite{scholler2020constant,becker2018red}, we provide the encoder network with relative displacements between subsequent observations, and not with the absolute coordinates of previous positions; \textit{ii)} regarding the design of the encoder network, it could be any module that extracts temporal features (\eg, Gated Recurrent Units (GRU)~\cite{cho2014learning,scholler2021flomo} or Transformers~\cite{vaswani2017attention,monti2022many}). In this work, the layout of the context encoder is a subset of the Temporal Fusion Transformer (TFT)~\cite{lim2021temporal}, a well-established and flexible backbone for time-series analysis/forecasting. In particular, we started from the original architecture and discarded the decoding modules, employing only the layers needed to encode the previous time-steps (referred as \quotationmarks{\textit{past inputs}} in the original paper) of the track $T$.

\tit{Dependence on scene-related visual information}~Importantly, one of the main issues we aim to address is the (lack of) adaptation to the scene under consideration. In this respect, existing approaches devise the same aggregated cost function for all conditions, which we argue may clash with the conditions we expect in real-world settings. Indeed, since different scenarios may display substantial differences (\eg, night/day, camera orientation, moving/stationary camera, \etc), some costs should be accordingly weighted more.

To provide such a feature, we propose to further condition the estimated density $f(\cdot | \ T, \theta)$ on a visual representation of the whole current frame. In particular, we exploit the variety of MOTSynth~\cite{fabbri2021motsynth} (comprising more than five hundred scenarios) and encode each frame $x^{t}$ through the CLIP's~\cite{radford2021learning} visual encoder, thus profiting from its widely-known zero-shot capabilities. On top of the extracted representations, we run the k-means algorithm and split them into $\lvert C \lvert=16$ clusters, each of which represents an abstract hyper-scenario. We then introduce the index $\mathbf{\hat{c}}$ of the cluster as a further conditioning variable:
\begin{align}
    f \equiv f([\Delta_p,\Delta_{w,h},\Delta_d] \ | \ T, \ \mathbf{\hat{c}} \ ,\theta) \\
    \text{where} \ \mathbf{\hat{c}} = \operatorname{argmin}_{i=1,\dots,\lvert C \lvert} \ || \operatorname{CLIP}_{v}(x^{t}) - c_i||_2^2,
\end{align}
and $c_i$ are the $\lvert C \lvert$ centroids retrieved through the k-means pass. Such a formulation also allows inference on novel scenarios, unseen during the training stage. To practically condition the model, we simply extend the context encoder $\operatorname{e}_{\theta_l}(\cdot)$ to take an additional learnable embedding $\operatorname{emb}_l[\hat{c}]$ as input, s.t. $\operatorname{e}_{l} \equiv \operatorname{e}_{\theta_l}(T, \operatorname{emb}_l[\hat{c}])$. In practice, in light of the TFT~\cite{lim2021temporal} layout employed by our context encoder, it becomes natural to include scene embeddings $\operatorname{emb}_l[\hat{c}]$ as static covariates~\cite{lim2021temporal, wen2017multi} -- \ie, something holding time-independent information about the time-series. We kindly refer the reader to the original paper~\cite{lim2021temporal} for all the important details regarding how the TFT uses covariates to influence the forward pass of each layer.
\subsubsection{Normalization of the cost matrix}
\label{subsec:cost_normalization}
Once the density estimator $f(\cdot | \ T, \ \mathbf{\hat{c}} \ ,\theta)$ has been trained, we exploit its output to fill the cost matrix $\Phi(D_j, T_i)$. Following Bastani~\etal~\cite{bastani2021self}, we apply a further normalization step, defined as follows:
\begin{align}
    \Phi^{\text{row}} = \frac{e^{\nicefrac{\Phi(D_j, T_i)}{\sigma}}}{\sum_k e^{\nicefrac{\Phi(D_j, T_k)}{\sigma}}},
    & \qquad
    \Phi^{\text{col}} = \frac{e^{\nicefrac{\Phi(D_j, T_i)}{\sigma}}}{\sum_k e^{\nicefrac{\Phi(D_k, T_i)}{\sigma}}},
    \\
    \hat{\Phi}({D_j, T_i}) = \operatorname{min}(&\Phi^{\text{row}}({D_j, T_i}), \Phi^{\text{col}}({D_j, T_i})).
\end{align}
In practice, we compute softmax (smoothed through a temperature hyperparameter $\sigma$) along rows and columns of $\Phi$; afterward, we take the cell-wise minimum between the two cost matrices. We finally pass the normalized cost matrix $\hat{\Phi}$ to the Hungarian algorithm for solving the associations.
\section{Experiments}
\subsection{Datasets}\label{sec:dataset}
We evaluate the performance of our models on the MOTSynth~\cite{fabbri2021motsynth}, MOT17~\cite{milan2016mot16}, and MOT20~\cite{dendorfer2021motchallenge} benchmarks.

\tit{MOTSynth}~We train both our main models on MOTSynth~\cite{fabbri2021motsynth}, a large synthetic dataset designed for pedestrian detection, tracking, and segmentation in urban environments. The dataset was generated using the photorealistic video game Grand Theft Auto V and comprises $764$ full-HD videos $1800$ frames long recorded at $20$ frames per second. The dataset includes a range of challenging scenarios, displaying various weather conditions, lighting, viewpoints, and pedestrian densities. The authors split the dataset into $190$ test sequences and $574$ train sequences; to select the hyperparameters of our models, we extract a holdout set of $32$ sequences from the training set. To speed up the evaluation phase, we assess the tracking performance only on the first $600$ frames of each testing video.

After a careful analysis of the dataset, we found a previous data-cleaning stage to be crucial. Indeed, as the dataset was collected automatically, there are several unrealistic dynamics, such as ground-truth tracks related to hidden people (\eg, behind walls) for many seconds. To address it, during the tracking performance evaluation, we disable annotations for targets not visible for more than \num{60} frames. Eventually, we re-activate the pedestrians in case they come back into the scene. In addition, we disable the annotations for pedestrians whose distance from the camera exceeds a certain threshold (fixed at $70$ meters), as we observe that they would slow down the training of the distance estimator with no evident benefits.

To better investigate the comparison of different trackers, we split the sequences of test sets based on their difficulty: namely easy, moderate, and hard. We characterize the tracking complexity through the HOTA, IDF1, and ID switches (IDs) metrics of three trackers (\ie, SORT, Bytetrack, and OC-SORT) and applied \textit{k}-means clustering to create three clusters of sequences. By doing so, the easy subset contains \num{26} sequences, the moderate subset contains \num{67} sequences, and the hard subset contains \num{42} sequences.

\tit{MOT17 and MOT20}~We use the standard benchmarks MOT17~\cite{milan2016mot16} and MOT20~\cite{dendorfer2021motchallenge} to evaluate multiple object tracking algorithms in crowded real-world scenarios. More specifically, MOT17 comprises seven training and seven test sequences of real-world scenarios. Similarly, the MOT20 benchmark, which is the latest addition to the MOTChallenge~\cite{dendorfer2021motchallenge}, also features challenging scenes with high pedestrian density, different indoor and outdoor locations, and different times of the day. Following the evaluation protocol introduced in~\cite{zhang2022bytetrack}, we define the MOT17 validation set by retaining half of its training set.
\subsection{Evaluation metrics}
\label{sec:metrics}
\begin{table*}[t]
    \centering
    \rowcolors{2}{lightgray}{}
    \begin{tabular}{lcccccccc}
    \toprule
    & \multicolumn{2}{c}{\textcolor{blue}{\textbf{Easy}}} & \multicolumn{2}{c}{\textcolor{orange}{\textbf{Moderate}}} & \multicolumn{2}{c}{\textcolor{red}{\textbf{Hard}}} & \multicolumn{2}{c}{\textit{\textbf{All}}}\\
    \midrule
    \rowcolor{white}\textbf{Metrics} & HOTA $\uparrow$ & IDF1 $\uparrow$ & HOTA $\uparrow$ & IDF1 $\uparrow$ & HOTA $\uparrow$ & IDF1 $\uparrow$ & HOTA $\uparrow$ & IDF1 $\uparrow$ \\
    \midrule
    SORT~\cite{bewley2016simple} & \ensuremath{63.48} & \ensuremath{79.40} & \ensuremath{50.31} & \ensuremath{62.11} & \ensuremath{37.48} & \ensuremath{45.13} & \ensuremath{48.42} & \ensuremath{59.05} \\
    ~+~\methnam GT & \hotagain{4.37} & \idfgain{7.41} & \hotagain{5.33} & \idfgain{9.09} & \hotagain{6.54} & \idfgain{10.88} & \hotagain{5.49} & \idfgain{9.62} \\
    ~+~\methnam & \hotagain{0.31} & \idfgain{0.97} & \hotagain{0.81} & \idfgain{1.63} & \hotagain{0.74} & \idfgain{1.56} & \hotagain{0.54} & \idfgain{1.22} \\
    \midrule
    ByteTrack~\cite{zhang2022bytetrack} & \ensuremath{63.22} & \ensuremath{80.84} & \ensuremath{49.91} & \ensuremath{62.46} & \ensuremath{37.61} & \ensuremath{46.15} & \ensuremath{48.21} & \ensuremath{59.79} \\
    ~+~\methnam GT & \hotagain{3.76} & \idfgain{2.82} & \hotagain{5.47} & \idfgain{5.51} & \hotagain{5.08} & \idfgain{4.60} & \hotagain{4.75} & \idfgain{4.54} \\
    ~+~\methnam & \hotagain{0.13} & \idfgain{1.80} & \hotagain{0.47} & \idfgain{1.21} & \hotagain{0.88} & \idfgain{1.81} & \hotagain{0.49} & \idfgain{1.41} \\
    \midrule
    OC-SORT~\cite{cao2022observation} & \ensuremath{65.56} & \ensuremath{81.61} & \ensuremath{52.42} & \ensuremath{63.50} & \ensuremath{38.10} & \ensuremath{45.48} & \ensuremath{49.96} & \ensuremath{60.16} \\
    ~+~\methnam GT & \hotagain{2.41} & \idfgain{3.76} & \hotagain{4.88} & \idfgain{7.70} & \hotagain{6.18} & \idfgain{9.55} & \hotagain{4.67} & \idfgain{7.67} \\
    ~+~\methnam & \hotagain{0.44} & \idfgain{0.84} & \hotagain{0.60} & \idfgain{1.09} & \hotagain{1.17} & \idfgain{1.96} & \hotagain{0.31} & \idfgain{0.70} \\
    \midrule
    Tracktor~\cite{bergmann2019tracking} & \ensuremath{46.59} & \ensuremath{49.40} & \ensuremath{29.15} & \ensuremath{28.81} & \ensuremath{21.58} & \ensuremath{22.68} & \ensuremath{30.82} & \ensuremath{30.91} \\
    \midrule
    \rowcolor{lightgray}CenterTrack~\cite{zhou2020tracking} & \ensuremath{43.75} & \ensuremath{43.01} & \ensuremath{28.13} & \ensuremath{24.05} & \ensuremath{20.03} & \ensuremath{18.79} & \ensuremath{29.32} & \ensuremath{26.20} \\
    \bottomrule
    \end{tabular}
    \vspace{1em}
    \caption{Tracking results on MOTSynth. For each tracker, we report its extended version using either our distance estimator (\ie, \methnam) and ground-truth distances, \ie, \methnam (GT). For a wider comparison, we also report two \textit{tracking-by-regression} approaches.}
    \label{tab:trackingresults}
\end{table*}
\tinytit{Tracking}~We adhere to the CLEAR metrics~\cite{bernardin2008evaluating} and provide the IDF1 and the higher-order tracking accuracy (HOTA)~\cite{luiten2021hota}. The former, the IDF1, evaluates the ability of the tracker to preserve the identities of objects through time and computes as the harmonic mean of the precision and recall of identity assignments between consecutive frames. The HOTA is a more comprehensive metric that simultaneously considers detection and association accuracy.

\tit{Distance estimation}~Besides considering the common metrics based on the squared error, we propose a novel measure, called Average Localization of Occluded objects Error (\textbf{ALOE}), tailored to measure the precision of the distance estimator for objects with a varying occlusion rate. In the following, we provide a summary of these metrics:
\begin{itemize}[leftmargin=5.5mm]
\setlength\itemsep{0em}
    \item \textbf{$\tau$-Accuracy}~\cite{ladicky2014pulling} (${\delta}_{<\tau}$): $\%$ of $\text{d}_i \ s.t. \  max(\frac{d_i}{d_i^*},\frac{d_i*}{d_i}) = \delta < \tau$ (\textit{e.g.}, $\tau=1.25$), represents the maximum allowed relative error;
    \item \textbf{Average Localization Precision}~\cite{xiang2015data,bertoni2019monoloco} (ALP${}_{@\tau}$): $\%$ of $d_i \ s.t. \  \lvert d_i - d_i^* \vert = \delta < \tau$ (\textit{e.g.}, $\tau \in \{0.5\text{m},1.\text{m},2.\text{m}\}$) is the mean average error in true distance range;
    \item \textbf{Error distances}~\cite{zhu2019learning}: absolute relative difference (Abs. Rel.), squared relative difference (Squa. Abs.), root mean squared error (RMSE), root mean squared error in the log-space (RMSE${}_{\text{log}}$);
    \item \textbf{ALOE${}_{[\tau_1:\tau_2]}$} -- Average Localization of Occluded objects Error (\textit{ours}): avg. absolute error (meters) for objects with an occlusion level between $\tau_1$ and $\tau_2$, with $\tau \in [0,1]$.
\end{itemize}
\subsection{Implementations details}
\label{sec:details}
\begin{table}[t]
    \centering
    \rowcolors{2}{lightgray}{}
    \setlength{\tabcolsep}{3pt}
    \begin{tabular}{lcccc}
    \toprule
    & \multicolumn{2}{c}{\textbf{MOT17}} & \multicolumn{2}{c}{\textbf{MOT20}} \\
    \midrule
    \rowcolor{white}\textbf{Metrics} & HOTA $\uparrow$ & IDF1 $\uparrow$ & HOTA $\uparrow$ & IDF1 $\uparrow$ \\
    \midrule
    SORT~\cite{bewley2016simple} & \ensuremath{64.17} & \ensuremath{72.98} & \ensuremath{60.56} & \ensuremath{74.30} \\
    ~+~\methnam & \hotagain{1.78} & \idfgain{1.41}  & \hotagain{0.15} & \idfgain{0.22} \\ 
    \midrule
    ByteTrack~\cite{zhang2022bytetrack} & \ensuremath{67.73} & \ensuremath{79.81} & \ensuremath{58.94} & \ensuremath{74.89} \\
    ~+~\methnam & \hotagain{0.40} & \idfgain{0.23}  & \hotagain{0.54} & \idfgain{0.06} \\ 
    \midrule
    OC-SORT~\cite{cao2022observation} & \ensuremath{66.22} & \ensuremath{77.74} & \ensuremath{55.18} & \ensuremath{71.22} \\
    ~+~\methnam & \hotagain{0.35} & \idfgain{1.12}  & \hotagain{0.53} & \idfgain{0.76} \\ 
    \midrule
    Tracktor++~\cite{bergmann2019tracking} & \ensuremath{44.66} & \ensuremath{55.00} & \ensuremath{30.36} & \ensuremath{40.63} \\
    \midrule
    \rowcolor{lightgray}CenterTrack~\cite{zhou2020tracking} & \ensuremath{48.59} & \ensuremath{58.44} & \ensuremath{31.69} & \ensuremath{41.43} \\
    \bottomrule
    \end{tabular}
    \vspace{1em}
    \caption{Tracking results on the validation set of MOT17 and the train set of MOT20~\cite{dendorfer2021motchallenge}.}
    \label{tab:trackingresultsmot}
\end{table}
We feed the distance estimator with video clips of $6$ frames, sampled with a uniform stride of length $8$: this way, each clip lasts approximately $2$ seconds. We adopt $1280\times720$ as input resolution, thus further preserving the visual cues. We set the batch size equal to $4$ and use Adam~\cite{kingma2014adam} as optimizer, with a learning rate of \num{5e-5}. On the other hand, the density estimator is trained with a batch size of $512$, Adam~\cite{kingma2014adam} as optimizer with learning rate \num{1e-3}. The normalizing flow consists of $16$ flow blocks, each comprising $64$ hidden neurons; regarding context conditioning, we fix the number of observed past observations $\lvert T \lvert=8$ and the number of visual clusters $C=16$. Unless otherwise specified, both the networks are trained only on synthetic data (\ie, the training set of MOTSynth); we leave the worth-noting investigation of possible transfer learning strategies for future works.
\subsection{Impact on tracking-by-detection}
\label{sec:tracking_evaluation}
\begin{table*}[t]
    \centering
    \rowcolors{2}{lightgray}{}
    \begin{tabular}{l@{\hskip 0.3in}c@{\hskip 0.3in}c@{\hskip 0.3in}ccc@{\hskip 0.3in}ccc}
    \toprule
     & & & \multicolumn{3}{c}{ALP $\uparrow$} & \multicolumn{3}{c}{ALOE $\downarrow$} \\
     \midrule
    \rowcolor{white}\textbf{Metrics} & $\delta_{<1.25}$ $\uparrow$ & RMSE $\downarrow$ & $@0.5m$ & $@1m$ & $@2m$ & ${}_{[0.3:0.5]}$ & ${}_{[0.5:0.75]}$ & ${}_{[0.75:1.]}$\\ 
    \midrule
    SVR & \ensuremath{26.7}\% & \ensuremath{12.5} & \ensuremath{3.4}\% & \ensuremath{6.8}\% & \ensuremath{13.8}\% & - & - & - \\
    DisNet~\cite{haseeb2018disnet} & \ensuremath{27.5}\% & \ensuremath{12.1} & \ensuremath{3.8}\% & \ensuremath{7.5}\% & \ensuremath{14.6}\% & - & - & - \\
    Zhu et al.~\cite{zhu2019learning} & \ensuremath{94.7}\% & \ensuremath{2.15} & \ensuremath{34.5}\% & \ensuremath{56.2}\% & \ensuremath{78.5}\% & \ensuremath{1.78} & \ensuremath{1.95} & \ensuremath{2.03} \\
    \midrule
    \textbf{DistSynth} & $\mathbf{\ensuremath{99.1}}$\% & $\mathbf{\ensuremath{1.91}}$ & $\mathbf{\ensuremath{48.0}}$\% & $\mathbf{\ensuremath{68.9}}$\% & $\mathbf{\ensuremath{86.1}}$\% & $\mathbf{\ensuremath{1.39}}$ & $\mathbf{\ensuremath{1.41}}$ & $\mathbf{\ensuremath{1.78}}$ \\
    \bottomrule
    \end{tabular}
    \vspace{1em}
    \caption{Comparison of various distance estimators on MOTSynth~\cite{fabbri2021motsynth}. Our \liftername exhibits superior performance across all the metrics reported. We highlight the enhancements observed in terms of ALOE, confirming an improved ability to withstand occlusions.}
    \label{tab:liftingresults}
\end{table*}
\begin{figure*}
\centering
\begin{tabular}[t]{cc}
\includegraphics[width=.45\linewidth,keepaspectratio]{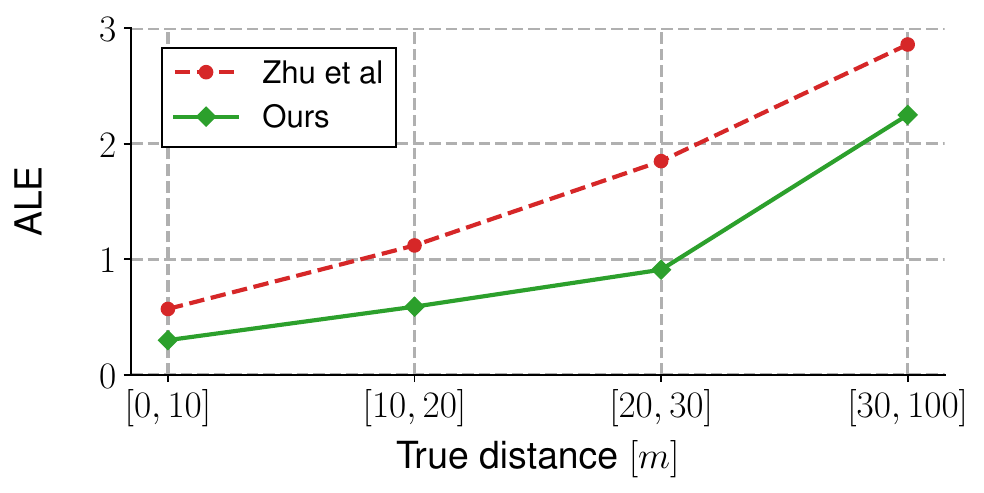} &
\includegraphics[width=.45\linewidth,keepaspectratio]{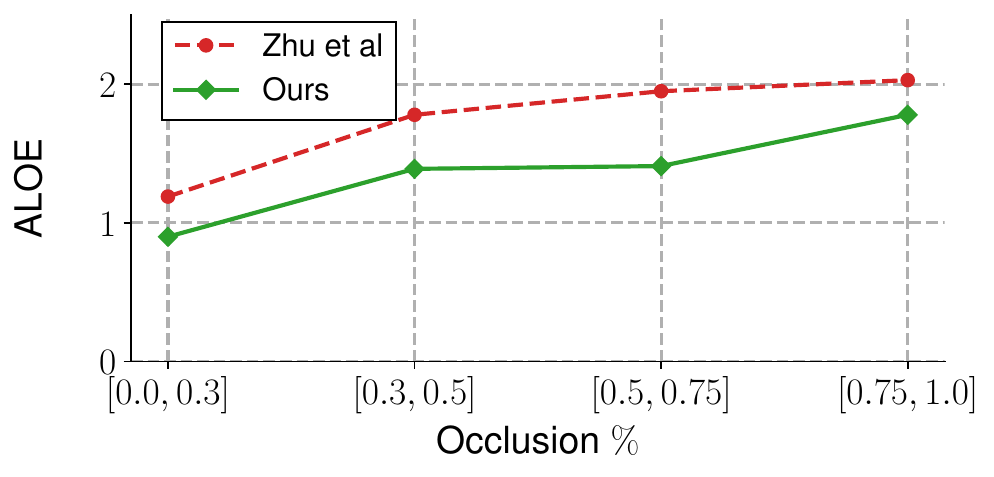} \\ 
(a) & (b)
\end{tabular}
\caption{
The ALE and ALOE metrics are evaluated on MOTSynth, and our method demonstrates significant improvements. Specifically, (a) our approach reduces ALE within the reported distance range shown in the plot, and (b) our method displays increased stability during occlusion events, resulting in superior performance, which can be attributed to our temporal approach.}
\label{fig:flatness}
\end{figure*}
In this section, we empirically show that our proposed method, applied to popular state-of-the-art \textit{tracking-by-detection} techniques, improves upon the MOTSynth and the MOTChallenge benchmarks (see~\cref{tab:trackingresults,tab:trackingresultsmot}).

On MOTSynth, we focus on three trackers (\ie, SORT~\cite{bewley2016simple}, Bytetrack~\cite{zhang2022bytetrack}, and OC-SORT~\cite{cao2022observation}) and adhere to the following common evaluation pipeline: \textit{i)} we compute predicted bounding boxes through YOLOX~\cite{ge2021yolox}; \textit{ii)} as our approach requires an estimate of per-pedestrian camera distances, we exploit YOLOX bounding boxes by providing them to the distance estimator \liftername (\cref{sec:distsec}); \textit{iii)} we finally integrate our density estimator \methnam into the pipeline of each tracker, applying the normalization described in~\cref{subsec:cost_normalization} before the Hungarian algorithm.

Additionally, we provide a further comparison -- termed \textbf{\methnam GT} (\ie, \textit{ground-truth}) -- that yields an upper bound for our approach. In practice, as standard \methnam, it relies on YOLOX detections to compute 2D displacements, but, differently, it leverages ground-truth distances (made available in MOTSynth) in place of the \liftername predictions. By doing so, it is possible to assess the potential of \methnam with near-perfect estimates.

We provide the results of such a comparison in~\cref{tab:trackingresults}. Our results indicate that \methnam enhances the performance of the considered trackers on all the MOTSynth splits, \ie, easy, moderate, hard, highlighting the benefits of our method across three levels of complexities. In particular, the improvements are of course huge when ground-truth distances are employed; nevertheless, a consistent gain can also be appreciated when leveraging estimated distances, leading to an improved identity accuracy reflected by a steady enhancement of the IDF1 metric.

As reported in~\cref{tab:trackingresultsmot}, the evaluation on the MOT17 and MOT20 benchmarks further shows that \methnam consistently improves the considered trackers in even more realistic scenarios (notably, SORT benefited the most from our approach). While evaluating on the MOT20 benchmark, we rely on the same YOLOX~\cite{ge2021yolox} model employed for MOT17. This particular YOLOX model was trained on two distinct datasets, namely CrowdHuman~\cite{shao2018crowdhuman} and the initial half of MOT17, which aligns with the training methodology adopted in ByteTrack~\cite{zhang2022bytetrack}.

As mentioned before, both \methnam and \liftername have been trained solely on synthetic data without additional fine-tuning, achieving still satisfying results on real data. Such a result opens the door to future research on how different components, such as the distance from the camera, can be used to advance multi-object tracking.
\subsection{Distance estimation: comparison with the State-of-the-art}
\label{sec:distance}
To assess the merits of the proposed distance estimator, we compare it with baselines and valid competitors from the current literature. We report the results of such a comparison in Tab.~\ref{tab:liftingresults} and refer the reader to the following paragraphs for a comprehensive analysis.

\tit{Comparison with Support Vector Regressor (SVR)}~It consists of a simple shallow baseline based on a support vector regressor, which exploits the dimensions of the bounding boxes (\ie, height and width). Through the comparison with such a naive approach, we would like to emphasize the gap \wrt the bias present in the task at hand \ie, the smaller the bounding box, the farther the pedestrian from the camera. As expected, the SVR approach yields low performance \wrt our method, due to its inability to generalize to objects with different aspect ratios.

\tit{Comparison with DisNet}~DisNet~\cite{haseeb2018disnet} consists of an MLP of 3 hidden layers, each of 100 hidden units with SeLU~\cite{klambauer2017self} activations. The network is fed with the relative width, height, and diagonal of bounding boxes, computed \wrt the image dimension; these features are then concatenated with three corresponding reference values (set to 175 cm, 55 cm, and 30 cm). As can be seen, the improvements of DisNet are marginal \wrt SVR, but its results are substantially lower than those obtained by both Zhu~\etal and our approach.
    
\tit{Comparison with Zhu \etal}~The model proposed by Zhu \etal~\cite{zhu2019learning} shares some similarities with our approach, as it relies on ResNet as feature extractor and RoI pooling to build pedestrian-level representations. However, thanks to the additional modules our model reckons on (\ie, the temporal module and the FPN branch), it is outperformed by our approach under all the considered metrics. Our advancements concerning ALE and ALOE, compared to Zhu \etal, are illustrated in~\cref{fig:flatness}.
\subsection{Analysis of \methnam}
\label{sec:density_analysis}
\begin{table}[t]
    \centering
    \rowcolors{2}{lightgray}{}
    \addtolength{\tabcolsep}{-4pt}
    \begin{tabular}{lcccccc}
    \toprule
    & & \textbf{MOTSynth} & \multicolumn{3}{c}{\textbf{MOT17}} \\
    \midrule
    & \textbf{cond.} & ${\mathbf{\operatorname{NLL}}}_{\downarrow}$ & ${\mathbf{\operatorname{NLL}}}_{\downarrow}$ & ${\text{HOTA}}_{\uparrow}$ & ${\text{IDF1}}_{\uparrow}$\\
    \midrule
    SORT~\cite{bewley2016simple} & - & - & - & \ensuremath{64.17} & \ensuremath{72.98} \\
    \midrule
    \midrule
    \methnam & \xmark & \ensuremath{-1.48} & \ensuremath{-5.66} & \ensuremath{65.34} & \ensuremath{74.77} \\
    \midrule
    \methnam & \cmark & $\mathbf{\ensuremath{-1.80}}$ & $\mathbf{\ensuremath{-5.81}}$ & $\mathbf{\ensuremath{65.95}}$ & $\mathbf{\ensuremath{75.71}}$ \\
    \midrule
    \midrule
    ${\text{\methnam}}_{\mathbf{\operatorname{FT}}}$ & \xmark & \ensuremath{-0.10} & \ensuremath{-7.29} & $\mathbf{\ensuremath{65.94}}$ & \ensuremath{75.97} \\
    \midrule
    ${\text{\methnam}}_{\mathbf{\operatorname{FT}}}$ & \cmark & $\mathbf{\ensuremath{-0.12}}$ & $\mathbf{\ensuremath{-7.50}}$ & \ensuremath{65.70} & $\mathbf{\ensuremath{76.22}}$ \\
    \bottomrule
    \end{tabular}
    \vspace{1em}
    \caption{For MOT17, ablative study W/o scenario-level conditioning (\ie, cond.) and W/o fine-tuning (\ie, ${\text{\methnam}}_{\mathbf{\operatorname{FT}}}$). Performance reported in terms of negative log-likelihood (${\mathbf{\operatorname{NLL}}}$) and HOTA/IDF1 for the evaluation of the resulting tracker.}
    \label{tab:nfablation}
\end{table}
We herein question the advantages of conditioning our density estimator on the scene under consideration. To do so, we focus on a single tracker (\ie, SORT) and compare how its tracking performance changes if the context encoder of \methnam (see ~\cref{sec:contextencoder}) considers only time-dependent information about the tracks and, hence, discards the scene-related visual information provided through cluster centroids $c_i$. From the results reported in~\cref{tab:nfablation} (second and third rows) it can be observed that visual conditioning (\ie, the row marked with \cmark) favorably leads to a lower negative log-likelihood on both the validation sets of MOTSynth and MOT17, as well as better HOTA and IDF1 results on MOT17. We interpret these findings as a confirmation of our conjectures about the advantages of designing a cost function that is aware of the scene.

Finally, we remind that only synthetic data have been used to train our models. However, it could be argued whether additional fine-tuning on real-world data could help. To shed light on this matter, we pick the best performing model attained on MOTSynth and carry out a final fine-tuning stage on the training set of MOT17, by training for further $20$ epochs with lowered learning rate. We report the performance of the resulting model (\ie, ${\text{\methnam}}_{\mathbf{\operatorname{FT}}}$) W/o visual conditioning. Two major findings emerge from an analysis of the last two rows of~\cref{tab:nfablation}: \textit{i)} as hold for frozen models, the introduction of visual cues leads to better results (with the only exception for the HOTA on MOT17); \textit{ii)} in general, additional training steps can profitably adapt \methnam to real-world scenarios, as confirmed by both the lower attained negative log-likelihood (equal to $-7.50$ after fine-tuning, in light of the value $-5.81$ prior fine-tuning) and higher tracking results.
\section{Conclusion}
This work presents a general approach for \textit{tracking-by-detection} algorithms, aimed to combine multi-modal costs into a single metric. To do so, it relies on a deep generative network, trained to approximate the conditional probability distribution of \textit{inlier costs} of correct associations. We prove the effectiveness of our approach by integrating 2D displacement and pedestrians' distances from the camera, delivered by a proposed spatio-temporal distance estimator, \liftername, designed for crowded in-the-wild scenarios. Remarkably, our method achieves competitive results on MOTSynth, MOT17, and MOT20 datasets. Notably, we show that training solely on synthetic data yields remarkable results, indicating the importance of simulated environments for future tracking applications, especially with non-collectible real-world annotations as 3D cues. We believe our work will drive further advancements toward the exploitation of 3D clues to enhance tracking approaches in crowded scenarios.
\section{Acknowledgement}
The research was financially supported by the Italian Ministry for University and Research -- through the PNRR project ECOSISTER ECS 00000033 CUP E93C22001100001 -- and the European Commission under the Next Generation EU programme PNRR - M4C2 - Investimento 1.3, Partenariato Esteso PE00000013 - \quotationmarks{FAIR - Future Artificial Intelligence Research} - Spoke 8 \quotationmarks{Pervasive AI}. Additionally, the research activities of Angelo Porrello have been partially supported by the Department of Engineering \quotationmarks{Enzo Ferrari} through the program FAR\_2023\_DIP -- CUP E93C23000280005. Finally, the PhD position of Gianluca Mancusi is partly financed by Tetra~Pak Packaging Solutions S.P.A., which also greatly supported the present research.

\end{document}